\pdfoutput=1
\documentclass[11pt,a4paper]{article}
\usepackage[hyperref]{acl2021}
\aclfinalcopy 
\usepackage{times}
\usepackage{nert}
\usepackage{latexsym}
\usepackage[T1]{fontenc}
\usepackage{bbm}
\usepackage{graphicx}
\usepackage{url}
\usepackage{amsmath}

\DeclareMathOperator*{\argmaxA}{arg\,max}

\usepackage{amsfonts}
\usepackage{xspace}
\usepackage{placeins}
\usepackage{soul}
\usepackage{caption}
\usepackage{newfloat}
\usepackage{tikz}
\definecolor{LimeGreen}{RGB}{179,253,148}
\definecolor{DarkLimeGreen}{RGB}{124,176,102}%{166,234,137}
\definecolor{DarkPink}{RGB}{255,135,135}
\definecolor{OliveGreen}{rgb}{0,0.6,0}

\usepackage{adjustbox}
\usepackage{array}
\usepackage{color, colortbl}
\newcolumntype{R}[2]{%
    >{\adjustbox{angle=#1,lap=\width-(#2)}\bgroup}%
    l%
    <{\egroup}%
}
\def\basiceval#1{\the\numexpr#1\relax}
\def\cca#1{\cellcolor{blue!#10}\ifnum #1>4\color{white!#10}\else\color{black!100}\fi{.#1}}
\def\ccb#1{\cellcolor{red!#10}\ifnum #1>4\color{white!#10}\else\color{black!100}\fi{-.#1}}
\def\ccaten#1{\cellcolor{blue!#1}\ifnum #1>40\color{white!100}\else\color{black!100}\fi{
\ifnum #1>10 .#1 \else .0#1 \fi
}}
\def\ccatentwo#1#2{\cellcolor{blue!\basiceval{#2*10}}\ifnum #2>4\color{white!#1}\else\color{black!100}\fi{#1.#2}}
\def\ccbten#1{\cellcolor{red!#1}\ifnum #1>40\color{white!#1}\else\color{black!100}\fi{
\ifnum #1>10 -.#1 \else -.0#1 \fi
}}
\def\ccbtentwo#1#2{\cellcolor{red!\basiceval{#2*5}}\ifnum #2>4\color{white!#1}\else\color{black!100}\fi{-#1.#2}}
\def\ccd#1{\cellcolor{blue!\basiceval{#1 * 2}}\ifnum #1>20\color{white!#1}\else\color{black!100}\fi{
\ifnum #1>10 .#1 \else .0#1 \fi
}}
\def\ccy#1{\cellcolor{yellow!\basiceval{#1 * 3}}\ifnum #1>50\color{white!#1}\else\color{black!100}\fi{
\ifnum #1>9 .#1 \else .0#1 \fi
}}

\definecolor{darkgreen}{HTML}{38761d}
\definecolor{darkyellow}{HTML}{ffa500}

\usepackage{microtype}

\title{\textit{NewsEdits}: A Dataset of Revision Histories for News Articles \\(Technical Report: Data Processing)}

\author{Alexander Spangher and Jonathan May \\ 
Information Sciences Institute \\
  University of Southern California \\ 
  \texttt{spangher@usc.edu}, \texttt{jonmay@isi.edu} \\}

\begin{document}
\maketitle
\begin{abstract}
News article revision histories have the potential to give us novel insights across varied fields of linguistics and social sciences. In this work, we present, to our knowledge, the first publicly available dataset of news article revision histories, or \textit{NewsEdits}. 

Our dataset is multilingual; it contains 1,278,804 articles with 4,609,430 versions from over 22 English- and French-language newspaper sources based in three countries. Across version pairs, we count 10.9 million %10,929,051 
added sentences; 8.9 million %8,908,254 
changed sentences and 6.8 million %6,807,624 
removed sentences. Within the changed sentences, we derive 72 million atomic edits. \textit{NewsEdits} is, to our knowledge, the largest corpus of revision histories of any domain.
\end{abstract}

\section{Introduction}

Revision histories gathered from various natural language domains like Wikipedia \cite{grundkiewicz2014wiked}, Wikihow \cite{faruqui2018wikiatomicedits} and student learner essays \cite{zhang2015annotation} have been studied widely in NLP. These corpora have been used for as varied tasks as language modeling \cite{yin2018learning}, sentence modification \cite{shah2020automatic}, argumentation design \cite{afrin2020annotation}, and even group collaboration dynamics \cite{muric2019collaboration}. %Because news article revisions histories have the potential to give us insights into text generation, event extraction and other various fields in NLP. 

By utilizing a novel domain for revision histories, \textit{news article revision histories}, we see the potential for strengthening methods for these aforementioned tasks, as well as the introduction of a novel set of tasks. Because news covers events (or world-states) that are constantly updating, we hypothesize that many edits in news either (1) incorporate new information (2) update events or (3) broaden perspectives.\footnote{This is in contrast with the distribution of edits in other domains like Wikipedia \cite{yang2017identifying}, which tend to focus on counter vandalism and syntax edits} Each of these categories of edits poses new questions that news edits are uniquely positioned to answer: What information is likely to change in the current state of a document? What must be incorporated? What perspectives need to be layered in, and when?

We offer, to our knowledge, the first publicly available corpus of news article revision histories, called \textit{NewsEdits}. We compile our corpus from various subcorpora collected by internet activists who track news article version changes. Each subcorpora is collected by monitoring article URLs, and downloading article text when a new version of the same news article is published.\footnote{As such, our dataset only contains differences between published versions of news articles, not intermediate drafts.}

Our dataset consists of 1,278,804 articles and 4,609,430 versions from over 22 English-language newspaper outlets. These outlets are based in the U.S., Great Britain and Canada. As in \newcite{tamori2017analyzing}, we compare article versions to each other on the sentence-level, and then, for matched sentences, the word level. We count 10,929,051 added sentences; 8,908,254 changed sentences and 6,807,624 removed sentences.% We additionally track XXX changed headlines.

Our contributions are the following:
\begin{enumerate}
    \item We introduce \textit{NewsEdits}, the first publicly available academic corpus of news article version histories, as well as, to our knowledge, the largest version-history corpus.
    \item We process the dataset to identify various forms of structural edits, in order to facilitate a wide range of possible analyses. We provide simple, lightweight visualization tools to render article-comparisons in an intuitive format.
\end{enumerate}

In the remainder of the paper, we start by comparing this work to other edit corpora. We then discuss the dataset curation and processing. In upcoming work, we will present a schema for characterizing edits in news articles, developed with journalists and graduate students in the communications field.

% related work table
\begin{table*}[t]
    \centering
    \begin{tabular}{|p{2cm}|p{4cm}|p{1.5cm}|p{2cm}|p{3cm}|}
    \hline
        Corpus & \# Revisions & Language & Source & Goal \\
    \hline
    \hline
        WiKed Error Corpus% \footnote{\cite{grundkiewicz2014wiked}} 
        &  12 million changed sentences 
        & English
        & Wikipedia
        & Grammatical Error Correction (GEC) \\
        \hline
        WikiAtomic-Edits% \footnote{\cite{faruqui2018wikiatomicedits}} 
        & 43 million ``atomic edits''\footnote{An ``atomic edit''}
        & 8 languages 
        & Wikipedia
        & Language Modeling \\
        \hline
        WiCoPaCo
        & 70,000 changed sentences
        & French
        & Wikipedia
        & GEC and Sentence paraphrasing \\
        \hline
        WikiHow-ToImprove
        & 2.7 million changed sentences
        & English
        & WikiHow
        & Version prediction, article improvement \\
        \hline
        NewsEdits
        & 8.9 million changed sentences, 10.9 million added sentences, 6.7 million removed sentences. 72 million atomic edits.
        & English and French
        & 22 media outlets
        & Language modeling, event sequencing, computational journalism \\
    \hline
    \end{tabular}
    \caption{A comparison of natural langauge revision history corpora.}
    \label{tab:my_label}
\end{table*}

% corpus statistics table
\begin{table*}[t]
\centering
\begin{tabular}{|l|r|r|l|l|l|l|l|}
\hline
Source & \# Articles & \# Versions &  Start & End & Ctry. & Lang.  & Coll. \\
\hline
    BBC &       307,616 & 1,244,490 & 2006-08  &  2021-01 &    U.K. &  En. & NS \\
    Guardian &  231,252 &   852,324 &  2012-01 &  2021-01 &    U.K. &  En. & NS \\
    Nytimes &   87,556 &    395,643 &  2012-08 &  2020-12 &    U.S. &  En. & NS \\
    Telegraph & 78,619 &    124,128 &  2017-01 &  2018-09 &    U.K. &  En. & NS \\
    Fox &       78,566 &    117,171 &  2017-01 &  2018-09 &    U.S. &  En. & DE \\
    CNN &       58,569 &    117,202 &  2017-01 &  2018-09 &    U.S. &  En. & DE \\
    Independent & 55,009 &  158,881 &  2014-01 &  2018-05 &    U.K. &  En. & NS \\
    CBC &         54,012 &  387,292 &  2017-08 &  2018-09 &  Ca. &  En. & DE \\
    Dailymail &   50,639 &  166,260 &  2017-01 &  2018-09 &    U.K. &  En. & DE \\
    BBC &         42,797 &  99,082 &  2017-01 &  2018-09 &     U.K. &  En. & DE \\
    La Presse &    40,978 &  73,447 &  2017-08 &  2018-09 &  Ca. & Fr-Ca. & DE \\
    Torontostar & 33,523 &  310,112 &  2017-08 &  2018-07 &    Ca. & En. &  DE \\
    Globemail &   32,552 &  91,820 &  2017-08 &  2018-09 &     Ca. & En. &  DE \\
    Reuters &     31,359 &  143,303 &  2017-01 &  2018-09 &    U.K. &  En. & DE \\
    National Post & 22,934 & 63,085 &  2017-08 &  2018-09 &    Ca. &  En. & DE \\
    Associated Press & 22,381 & 97,314 &  2017-01 &  2018-09 &  U.S. & En. & DE \\
    Washington Post &  19,184 & 68,612 &  2014-01 &  2020-07 &   U.S. & En. &  NS \\
    Toronto Sun &      19,121 & 46,353 &  2017-08 &  2018-09 &   Ca. &  En. & DE \\
    Calgary Herald &   7,728 &  33,427 &  2017-08 &  2018-09 &   Ca. & En. &  DE \\
    The Rebel &        4,344 &  19,383 &  2017-08 &  2018-09 &  Ca. &  En. &  DE \\
    Canada Land &      65 &     101 &  2017-12 &  2018-09 &    Ca. &  En. & DE \\
\hline
\end{tabular}
\caption{A summary of the number of total number of articles and versions for different media outlets which comprise our dataset. Also shown is the original collection that they were derived from (DE for DiffEngine, and NS from NewsSniffer), and the date-ranges during which articles from each outlet were collected.}
\label{tbl:source_list}
\end{table*}

\section{Related Work}

Previous work in natural language revision histories has primarily focused on two primary domains: student learner essays and Wikipedia. There is, additionally, emerging work in using WikiHow for similar ends as Wikipedia \cite{anthonio2020wikihowtoimprove,bhat2020towards}.

\subsection{Wikipedia Revisions}

Wikipedia is a resource that is often used in text-revision research. %It is the source of the largest revision-histories corpora to date.
Many tasks have benefited from studying Wikipedia revisions, such as text simplification \cite{yatskar2010sake}, textual entailment \cite{zanzotto2010expanding} and discourse learning in edits \cite{daxenberger2012corpus, daxenberger2013automatically, fong2010did}. Discourse learning for edits, as in other branches of discourse learning, focuses on developing schemas to elucidate the function or purpose of each edit, and then performing classification on these schema \cite{faigley1981analyzing}. One such work, by \newcite{yang2017identifying}, developed a schema for Wikipedia edit intentions, %\footnote{
% Interestingly, the schema development was done publicly, in collaboration with veteran Wikipedia editors on a Wikipedia thread, documented here: \url{https://en.wikipedia.org/wiki/Wikipedia_talk:Labels/Edit_types/Taxonomy}} 
included Wikepedia-specific edit categories, such as \textit{Counter-Vandalism} and \textit{Wikification}, as well as general categories like \textit{Elaboration} and \textit{Refactoring}. We take the general categories as a starting point for our own work.

The two largest-scale corpora to be processed and released, to our knoweldge, are the WikEd Error Corpus, in English, which contains 12 million sentences and 14 million revisions \cite{grundkiewicz2014wiked}, and WikiAtomicEdits, which contains 43 million ``atomic edits''.\footnote{Authors are unclear about how many sentences this corresponds to, but in our work we found, on average, roughly $4$ atomic edits per changed sentence} The WikEd Error Corpus has been used primarily for Grammar Error Correction (GEC), while WikiAtomicEdits has been used primarily for language modeling \cite{yin2018learning, shah2020automatic}.

While our work has roughly the same number of changed sentences as the Wikipedia-based corpora (8.9 million), we have roughly twice as many atomic edits (72 million), and added/removed sentences, which neither Wikipedia corpus reports. 

\subsection{Student Learner Essays}

Student Learner essays are another area of focus in revisions research. Such research focuses on editing revisions made during student essay-writing, particularly focusing on non-English speakers. Because of the difficulties in collecting data in this domain, most natural datasets tend to be small \cite{leacock2010automated}. In recent work, researchers create an editing platform; they instruct students to write 3 drafts of an essay using their platform, and gather revisions made during the writing process \cite{zhang2017corpus}. They collect 60 essays with 180 versions, or 3 drafts per essay, and they focus on classifying the discursive purpose of each edit (i.e. what the edit introduces, like \textit{Evidence}, or \textit{Claims}). 

In this vein, researchers have constructed Automated Writing Evaluation systems (AWE) and used these systems to evaluate writing when a student submits a draft \cite{wang2020erevis,zhang2020engaging, zhang2015annotation}. In one recent work by \newcite{afrin2020annotation}, researchers compile 143 essays with 286 versions, or 2 versions each. They develop an annotation scheme focused on improving the writing (ex: \textit{Evidence: Relevant} or \textit{Irrelevant}). 

While such work is interesting because the generation process is fully controlled: i.e., students can be instructed to write about similar topics and be given similar feedback between rounds, the corpora collected are small by modern machine learning standards.

\subsection{News}
\subsubsection{Academic Work}
Despite the centrality of published news articles as sources to some of the most fundamental corpora in NLP \cite{marcus1993building, carlson2003building, pustejovsky2003timebank, ace2005}, there is, to our knowledge, no revision-histories corpus based on news edits currently published in the academic literature. For research on linguistics, corpora from multiple domains can capture relevant mechanisms, but for research on language describing real-world events -- such as event extraction or temporality -- news is still the gold standard.

We are aware of just one academic work, and its followup, that focuses on news edits \cite{tamori2017analyzing, hitomi2017proofread}. Authors analyzed news edits to predict quality of news articles, as well as to predict editing operations performed during text generation. A dataset of news revisions from a Japanese newspaper was used, but not released publicly.\footnote{Dataset could not be released due to copyright infringement, according to the authors in response to inquiry.} 

In our work, in contrast, we publicly release a large dataset and outline several tasks that, in our opinion, are tasks for which news corpora are best suited. In previously studied revision-corpora, researchers typically assume, implicitly, that the writing focuses on a relatively static world-state -- both Wikipedia and student learner essays tend to be focused on historical events or arguments \cite{yang2017identifying}. Thus in previous corpora, the nature of edits are primarily argumentative or corrective. However, news articles very often cover updating events. This difference has important implications for the kinds of edits we expect in our corpora.

We are not aware of any work using WikiNews\footnote{\url{https://en.wikinews.org/wiki/Main_Page}} as a source of revision histories, despite its proximity, as a domain, to other corpora that have been used in revision-history research, like Wikipedia and WikiHow. WikiNews corpora have been used for study in the communications literature \cite{thorsen2008journalistic, roessing2019wikis}, as well as in subfields of NLP, including document summarization \cite{bravo2012zipf}, timeline synthesis \cite{zhang2017towards, minard2016meantime} and word-identification tasks \cite{yimam2017cwig3g2}. While we are aware of one work, on identifying entity-salience, that uses WikiNews editor-annotations as part of their task \cite{wu2020wn}, we are not aware of any other works that utilize the wiki structure of WikiNews, including its revision histories. We considered using WikiNews as an additional source, but were concerned that its generation process (i.e. as a community-generated resource) was significantly different than the other sources we compiled (i.e. professionally developed articles). In future work, when we are better able to do more extensive comparison, we will consider include it in our collection of news revision corpora. %for the present work, decided that its emphasis on secondary sources might  it against tasks that we were interested in.

\subsubsection{Nonacademic Work}
Since at least 2006, internet activists have tracked changes made to major digital news articles \cite{bbc2006}. In 2012 \url{NewsDiffs.org} became one of the first websites to gain popular media attention for tracking changes to articles published in the \textit{New York Times}, CNN, \textit{Politico}, and others \cite{brisbane_2012, burke_2016, jones2017using, fass2014revealing}. Other similar trackers have since (or concurrently) emerged, such as NewsSniffer\footnote{\url{https://www.newssniffer.co.uk/}} and DiffEngine.\footnote{\url{https://github.com/DocNow/diffengine}} These tools collect article versions from RSS feeds and the Internet Archive. Using open-sourced technology, dozens of major newspapers are being analyzed,\footnote{\url{https://twitter.com/i/lists/821699483088076802}} as well as thousands of government websites\footnote{\url{https://envirodatagov.org/federal-environmental-web-tracker-about-page/}}. Such diffs have been used by journalists, communications scholars and media critics to study instances of gender and racial bias in earlier article drafts,\footnote{\url{http://www.newsdiffs.org/diff/192021/192137/www.nytimes.com/2013/03/31/science/space/yvonne-brill-rocket-scientist-dies-at-88.html}} shifting portrayals of social events, \cite{johnson2016effect}, and lack of media transparency \cite{gourarie_2015}. We utilize this work in the construction of our corpora, and are actively exploring efforts in this space for additional sources of data and analysis.

% db schemas
\begin{table*}[t]
\centering
\subfloat[DB schema for the \texttt{article} table. \texttt{SOURCE}, \texttt{A\_ID} and \texttt{VERSION\_ID} are the primary key columns. \label{tbl:layout_articles}]{
\begin{tabular}{|p{2.9cm}|p{1cm}||p{2.9cm}|p{1cm}||p{2.9cm}|p{1cm}|}
\hline
         Column Name &     Type  & Column Name &     Type  &  Column Name &     Type  \\
\hline
    SOURCE & index &      TITLE & text & CREATED & text \\
    A\_ID &  index &      URL &  text &  ARCHIVE\_URL &   text \\
    VERSION\_ID & index & TEXT & text &  NUM\_VERSIONS &  int \\
\hline
\end{tabular}
} \\
\subfloat[DB schema for the \texttt{sentence\_diffs} table (\texttt{word\_diffs} is similar). Table compares \textit{version pairs} of articles. The rows in the table are on the sentence-level; \texttt{V\_OLD\_ID} refers to the index of the old version, \texttt{V\_NEW\_ID} refers to the index of the new version. \texttt{TAG\_OLD} gives information for how to transition from the old version to the new version; \texttt{TAG\_NEW} is the inverse.\label{tbl:layout_sdiffs}]{
\begin{tabular}{|p{2.9cm}|p{1cm}||p{2.9cm}|p{1cm}||p{2.9cm}|p{1cm}|}
\hline
   Column Name &     Type  & Column Name &     Type  &  Column Name &     Type  \\
\hline
    SOURCE &     index & V\_NEW\_ID &    index &  TAG\_OLD &     text \\
    A\_ID &      index & SENTENCE\_ID &  index &  SENT\_NEW &     text \\
    V\_OLD\_ID & index & SENT\_OLD &     text &   TAG\_NEW &     text \\
\hline
\end{tabular}}
\caption{Schemas for two databases central to our content organization scheme.}
\label{tbl:dbschemas}
\end{table*}

% demo table
\begin{table*}
\small
\centering
% table 1 
\subfloat[
    Demo 1: Word-Level atomic edit corrections applied when a sentence-level match is found, using the \texttt{difflib} Python library.
    \label{tbl:demo2}
]{
\begin{tabular}{|p{.4cm}|p{.4cm}|p{5.6cm}|p{5.6cm}|p{.4cm}|} 
\hline
Sent Idx & Old Tag & Old Version & New Version & New Tag\\
\hline
1 & M 1 C & 
\cellcolor{pink}The Bundesbank would only refer to an interview \colorbox{DarkPink}{Mr.} \colorbox{DarkPink}{Weidmann} \colorbox{DarkPink}{gave} \colorbox{DarkPink}{to} Der Spiegel magazine last week, in which \colorbox{DarkPink}{he} said, ``I can \colorbox{DarkPink}{do} my \colorbox{DarkPink}{job} best \colorbox{DarkPink}{by} \colorbox{DarkPink}{staying} in office.'' &  
\cellcolor{LimeGreen}The Bundesbank would only refer to an interview \colorbox{DarkLimeGreen}{published} \colorbox{DarkLimeGreen}{in} Der Spiegel magazine last week, in which \colorbox{DarkLimeGreen}{Mr.} \colorbox{DarkLimeGreen}{Weidmann} said, ``I can \colorbox{DarkLimeGreen}{carry} \colorbox{DarkLimeGreen}{out} my \colorbox{DarkLimeGreen}{duty} best \colorbox{DarkLimeGreen}{if} \colorbox{DarkLimeGreen}{I} \colorbox{DarkLimeGreen}{remain} in office.'' & 
M 1 C \\
\hline
\end{tabular}
}
\\
\vspace{.1cm}
% table 2
\subfloat[
    Demo 2: A sentence that is split results in the addition of a new sentence, but is matched with the previous dependent clause. Minimal word-level edits are applied.
    \label{tbl:demo1}
]{
\small
\begin{tabular}{|p{.4cm}|p{.4cm}|p{5.6cm}|p{5.6cm}|p{.4cm}|} 
\hline
Sent Idx & Old Tag & Old Version & New Version & New Tag\\
\hline
1 & M 1 2 C & 
\cellcolor{pink}DALLAS—Ebola patient Thomas Eric Duncan told his fiancee the day he was diagnosed last week that he regrets exposing her to the deadly virus \colorbox{DarkPink}{and} \colorbox{DarkPink}{had} he known he was carrying Ebola, he would have “preferred to stay in Liberia and died than bring this to you,” a family friend said & 
\cellcolor{LimeGreen} DALLAS—Ebola patient Thomas Eric Duncan told his fiancee the day he was diagnosed last week that he regrets exposing her to the deadly virus\colorbox{DarkLimeGreen}{.}  & 
M 1 \\
\hline
2 & {} & {} & \cellcolor{LimeGreen} \colorbox{DarkLimeGreen}{Had} he known he was carrying Ebola, he would have “preferred to stay in Liberia and died than bring this to you,” a family friend said. & 
M 1 C \\
\hline
\end{tabular}
}
\\
\vspace{.1cm}
% table 3
\subfloat[Demo 3: Two features shown: (1) Refactoring, or order-swapping, makes sentences appear as though they have been deleted and then added. Swapped sentences are matched through their tags. (2) The last sentence is a newly added sentence and is not matched with any other sentence.\label{tbl:demo3}]{
\begin{tabular}{|p{.4cm}|p{.4cm}|p{5.6cm}|p{5.6cm}|p{.4cm}|} 
\hline
Sent Idx & Old Tag & Old Version & New Version & New Tag\\
\hline
1 & M 2 U & \cellcolor{pink} ``The mother, this was the first time seeing her son since he got to the States. & 
\cellcolor{LimeGreen} ``She has not seen him for 12 years, and the first time she saw him was through a monitor,'' said Lloyd. & M 2 U \\
\hline
2 & M 1 U & \cellcolor{pink} She has not seen him for 12 years, and the first time she saw him was through a monitor,'' said Lloyd. &
\cellcolor{LimeGreen} ``The mother, this was the first time seeing her son since he got to the States.''
& M 1 U \\
\hline
3 & {} & {} &
\cellcolor{LimeGreen} ``She wept, and wept, and wept.''
& A \\
\hline
\end{tabular}
}
\vspace{.1cm}
\caption{Here we show demos of three tricky edge-cases and how our tagging scheme handles them. 
\texttt{Old Tag} annotates a \texttt{Old Version} relative to changes in the \texttt{New Version} (or ``converts'' the \texttt{Old Version} to the \texttt{New Version}). \texttt{New Tag} is the inverse. Tag components: \textbf{Component 1: M, A, R.} Whether the sentence is \textbf{M}atched, \textbf{A}dded, or \textbf{R}emoved. \textbf{Component 2: Index.} If \textbf{M}atched, what is the index of the sentence in version that it is matched to. \textbf{Component 3: C, U.} If \textbf{M}atched, is the sentence \textbf{C}hanged or \textbf{U}nchanged.}
\label{tbl:demos}
\end{table*}

\section{Dataset}

We combine data from two primary sources: NewsSniffer (NS) and Twitter accounts powered by DiffEngine (DE). Currently, we have gathered data for 22 media outlets. The number of articles per outlet, as well as the time-frames for which we have collected data, is listed in Table \ref{tbl:source_list}. We are in the process of accumulating more data from \textit{La Nacion}, \textit{Clarìn}, \textit{Le Soir}, and \textit{Le Monde} from internet activists to broaden the number of languages we currently have access to.

\subsection{Dataset Tables and Fields}

Our dataset is released in a set of 5 SQLite tables. Three of them are primary data tables, and two are summary-statistic tables. Our primary data tables are: \texttt{articles}, \texttt{sentence\_diffs}, \texttt{word\_diffs}; the first two of which are shown in Tables \ref{tbl:layout_articles} and \ref{tbl:layout_sdiffs} (\texttt{word\_diffs} shares a similar structure with \texttt{sentence\_diffs}). We compile two summary statistics tables to cache statistics from \texttt{sentence\_diffs} and \texttt{word\_diffs}; they calculate metrics such as \texttt{NUM\_SENTENCES\_ADDED} and \texttt{NUM\_SENTENCES\_REMOVED} per article.\footnote{These summary statistic tables make it convenient to, say, filter \texttt{sentence\_diffs} in order train a model on all articles that have one sentence added; or all articles that have no sentences removed.}

The \texttt{sentence\_diffs} data table's schema is shown in Table \ref{tbl:dbschemas} and some column-abbreviated sample rows are shown in Table \ref{tbl:demos}. As can be seen, the diffs are calculated and organized on a sentence-level. Each row shows a comparison of sentences between \textit{two adjacent versions of the same article.}\footnote{So, for instance, article A, with versions 1, 2 where each version has sentences i, ii, iii, would have 3 rows (assuming sentences were similar): A.1-2.i, A.1-2.ii, A.1-2.iii.} Every row in \texttt{sentence\_diffs} contains index columns: \texttt{SOURCE}, \texttt{A\_ID}, \texttt{VERSION\_OLD}, and \texttt{VERSION\_NEW}. These columns can be used to uniquely map each row in \texttt{sentence\_diffs} to \textit{two} rows in \texttt{article}.\footnote{One mapping for \texttt{sentence\_diffs.VERSION\_OLD} = \texttt{article.VERSION\_ID} and one mapping for \texttt{sentence\_diffs.VERSION\_NEW} = \texttt{article.VERSION\_ID}.} 

\subsection{\texttt{TAG} columns in \texttt{sentence\_diffs}}

The columns \texttt{TAG\_OLD} and \texttt{TAG\_NEW} in \texttt{sentence\_diffs} have specific meaning: how to transform from version to its adjacent version. In other words, \texttt{TAG\_OLD} conveys where to find \texttt{SENT\_OLD} in \texttt{VERSION\_NEW} and whether to change it, whereas \texttt{TAG\_NEW} does the same for \texttt{SENT\_NEW} in \texttt{VERSION\_OLD}. 

More concretely, consider the examples in Table \ref{tbl:demo1}, \ref{tbl:demo2} and \ref{tbl:demo3}. As can be seen, each tag is 3-part and has the following components. \textbf{Component 1} can be either \textbf{M}, \textbf{A}, or \textbf{R}. \textbf{M} means that the sentence in the current version was \textbf{M}atched with a sentence in the adjacent version, \textbf{A} means that a sentence was \textbf{A}dded to the new version and \textbf{R} means the sentence was \textbf{R}emoved from the old version.\footnote{i.e. an \textbf{A}dded row is not present in the old version and a \textbf{R}emoved row is not present in the new version. They have essentially the same meaning and we could have condensed notation, but we felt this was more intuitive.} \textbf{Component 2} is only present for \textbf{M}atched sentences, and refers to the index or indices of the sentence(s) in the adjacent version\footnote{I.e. in \texttt{TAG\_OLD}, the index refers to the \texttt{SENTENCE\_ID} of \texttt{SENT\_NEW}}. Additionally, \textbf{Component 3} is also only present if the sentence is \textbf{M}atched. It can be either \textbf{C} or  \textbf{U}. \textbf{C} refers to whether the matched sentence was \textbf{C}hanged and \textbf{U} to whether it was \textbf{U}nchanged.

Although not shown or described in detail, all \textbf{M} sentences have corresponding entry-matches in \texttt{word\_diffs} table, which has a similar schema and tagging aim.

A user might use these tags in the following ways:

\begin{enumerate}
    \item \label{itm:firstusecase} To compare only atomic edits, as in \newcite{faruqui2018wikiatomicedits}, a user could filter \texttt{sentence\_diffs} to sentences where \textbf{M..C} is in \texttt{TAG\_OLD} (or equivalently, \texttt{TAG\_NEW}). Then, they would join \texttt{TAG\_OLD.Component\_2} with \texttt{SENTENCE\_ID}. Finally, they would select \texttt{SENT\_OLD}, \texttt{SENT\_NEW}.\footnote{or simply look in the \texttt{word\_diffs} table.}
    \item To view only refactorings, or when a sentence is moved from one location in the article to another, a user could filter \texttt{sentence\_diffs} to only sentences containing \textbf{M..U} and follow a similar join process as in use-case \ref{itm:firstusecase}.
    \item\label{itm:thirdusecase} To model which sentences might be added, i.e. $p(\text{sentence}_i \in \text{article}_{t+1} | \text{sentence}_i \nin \text{article}_t)$, a user would select all sentences in \texttt{SENT\_OLD}, and all sentences in \texttt{SENT\_NEW} where \textbf{A} is in \texttt{TAG\_NEW}.
    \item To model the inverse of use-case \ref{itm:thirdusecase}, i.e. which sentences would be removed, or $p(\text{sentence}_i \nin \text{article}_{t+1} | \text{sentence}_i \in \text{article}_t)$, a user would select all sentences in \texttt{SENT\_NEW}, and all sentences in \texttt{SENT\_OLD} where \textbf{R} is in \texttt{TAG\_OLD}.
\end{enumerate}

\subsection{Assigning \texttt{sentence\_diff} Tag Columns}

To assign the tags, we need to determine which sentences are \textbf{A}dded, \textbf{R}emoved and \textbf{M}atched. We seek to have our dataset reflect a general principle: sentences tagged as \textbf{A}dded should contain novel information and sentences tagged as \textbf{R}emoved should delete information that the journalist wishes to delete. Sentences tagged as \textbf{M}atched should be substantially similar except for syntactic changes, rephrasing, and updated information.

\subsubsection{Matching Algorithm}

To do this, we develop an asymmetrical sentence-matching algorithm. The examples shown in Tables \ref{tbl:demo1}, \ref{tbl:demo2} and \ref{tbl:demo3} illustrate our requirements. The first example, shown in Table \ref{tbl:demo1}, occurs when a sentence is edited syntactically, but its meaning does not change\footnote{Syntactic changes: synonyms are used, or phrasing is condensed, but substantially new information is not added}. So, we need our sentence-matching algorithm to use a sentence-similarity measure that considers semantic changes and does not consider surface-level changes. The second example, shown in Table \ref{tbl:demo2}, occurs when a sentence is split (or inversely, two sentences are merged.) Thus, we need our sentence matching algorithm to consider many-to-one matchings for sentences. The third example, shown in Table \ref{tbl:demo3}, occurs when sentence-order is rearranged, arbitrarily, throughout a piece. Finally, we need our sentence-matching algorithm to perform all pairwise comparisons of sentences. 

Our algorithm is given in Algorithm \ref{alg:matching}. Given a list of sentences for pairwise versions, our algorirthm computes the asymmetrical similarity between sentence-pairs in the Cartesian product of the sentences of adjacent article versions. It returns two mappings: (1) from sentences in the old version to the new, and (2) from sentences in the new to the old. This relies on an effective sentence-similarity score.

\begin{algorithm}[t]
\SetAlgoLined
\SetKwInOut{Input}{input}
\SetKwInOut{Output}{output}
\Input{Article versions $v_{old}$, $v_{new}$, Match Threshold $T$}
\Output{maps $m_{old \rightarrow new}$, $m_{old \leftarrow new}$}
initialize;\\
$m_{old \rightarrow new}$, $m_{old \leftarrow new}$ = \{\}, \{\};\\
 \tcp{match $v_{old} \rightarrow v_{new}$}
 \For{$(i, s_i) \in v_{old}$}{
    $d = \max_{s_j \in v_{new}} \text{Sim}_{asym}(s_i, s_j)$\\
    $j = \argmaxA_{s_j \in v_{new}} \text{Sim}_{asym}(s_i, s_j)$\\
    $m_{old \rightarrow new}\left[i\right] = j \times \mathbbm{1} \left[d > T\right]$\\
 }
 \tcp{match $v_{old} \leftarrow v_{new}$}
 \For{$(j, s_j) \in v_{new}$}{
    $d = \max_{s_i \in v_{old}} \text{Sim}_{asym}(s_j, s_i)$\\
    $i = \argmaxA_{s_i \in v_{old}} \text{Sim}_{asym}(s_j, s_i)$\\
        $m_{old \leftarrow new}\left[j\right] = i \times \mathbbm{1} \left[d > T\right]$
    
 }
\caption{Asymmetrical sentence-matching algorithm. Input $v_{old}$, $v_{new}$ are lists of sentences, and output is an index mapper. If a sentence maps to 0 (i.e. $d < T$), there is no match. $Sim_{asym}$ is described in text.}
\label{alg:matching}
\end{algorithm}

\subsubsection{Sentence Matching}

There is a wide body of research in sentence-matching \cite{quan2019efficient, abujar2019sentence, yao2018novel, chen2018sentence} including BERT-based approaches \cite{reimers2019sentence}, dependency-tree approaches \cite{le2018acv} and \cite{allan2003retrieval, achananuparp2008evaluation}. We desire a measure that considers semantics over syntactical changes, yet can appropriately match named entities (names, places or organizations) or other specific words not likely to be found in pretrained models.

While we are still evaluating the effectiveness of several methods, our primary method so far is based on a matching algorithm similar to \textit{maximum alignment} method, described by \newcite{kajiwara2016building}, shown below, where $\phi(x_i, y_j) :=$ similarity between word $x_i$ and $y_j$. 

\begin{align*}
\text{Sim}_{asym}(x, y) &= \frac{1}{|x|}\sum_{i=1}^{|x|}\max_j \phi(x_i, y_j)
\end{align*}

This approach allows us to identify unidirectional matches, where sentence \textit{a} might be a subsentence of sentence \textit{b}, as shown in Table~\ref{tbl:demo2}. We use this method because it is similar to one used in prior work on news article revision histories \cite{tamori2017analyzing}. %To do so, we use $Sim_{asym}$ and find all matches from $x \rightarrow y$ and $x \leftarrow y$.

We test several word-similarity functions, $\phi$. The first uses a simple lexical overlap, where $\phi(x, y) = 1$ if $lemma(x) = lemma(y)$ and 0 otherwise. This is based off of the hypothesis that proper nouns make up the majority of important components in our desired similarity measure, and these are poorly captured by large language models. The second uses contextual word-embeddings, where $\phi(x, y) = Emb(x) \cdot Emb(y)$, and $Emb(x)$ is the word-embeddings derived from a pretrained language model (Albert-XXLarge-Uncased.)\cite{lan2019albert}. We are still testing different similarity thresholds (T in Algorithm \ref{alg:matching}) and still determining how well this embedding function captures proper nouns and entities.

\section{Discussion and Possible Tasks}
\label{sct:discussion}
In this work, we have introduced, to our knowledge, the largest dataset of revision histories ever, and the first public dataset of news revision histories into in the academic literature.

\subsection{Tasks}
We hope this resource will prove useful as a domain that is far more general in terms of standards and style than domains such as Wikipedia and WikiHow, for which revisions data already exists. Thus, tasks based off these existing corpora, such as edit language modeling \cite{yin2018learning} and fact-guided revisions \cite{shah2020automatic}, should stand to benefit.

However, as mentioned previously, news articles are, more often than Wikipedia or Student Learner articles, based on a world-state that is dynamically changing. Thus, we expect that edits in news articles are more likely to describe events that are changing and include primary-source material (in contrast to another source we considered, WikiNews). This has important implications for linguistic inquiries, such as:

\begin{enumerate}
    \item Event-temporal relations, as edits update events that are changing through time \cite{ning2018multi}
    \item Headline Generation \cite{shen2017recent}
    \item Fact-guided updates \cite{shah2020automatic}
\end{enumerate}

This dataset is also interesting as it captures the lifecycles of news articles, which have standard arcs. Breaking news typically gets published first as a bare-bones paragraph containing a main event. Within the next few hours and days, additional quotes, contextual and explainer-paragraphs are added until the article comes to resemble a more standard news article.\footnote{Much of our insight in this section is derived from our own professional experience working in newsrooms as well as journalists we interviewed before and during the writing of this article.} Thus, there are tasks this corpus sheds light on that have important implications for computational journalism:

\begin{enumerate}
    \item News information gathering \cite{spangher2020sourcefinding}
    \item Contextualization: discourse and article structure \cite{choubey2020discourse, spangher2021multitask}
    \item How framing changes in response to an unfolding story \cite{spangher2021annenberg}
\end{enumerate}

Finally, the nature of our article updates -- each version is captured upon publication, not drafting -- also contain fewer grammatical and spelling errors than, say, student learner essays. Thus, edits made to change the syntax of a sentence rather than introduce or change information, are more likely to have stylistic purpose. This might lead to interesting subtasks in several fields of NLP, such as:

\begin{enumerate}
    \item Style transfer \cite{fu2018style}
    \item Bias in news articles \cite{mehrabi2020man}
    \item Cross-cultural sensitivity \cite{tian2020identifying}
\end{enumerate}

In short, there are many tasks that we see emerging from such a dataset, and several tasks that are currently ongoing.

\subsection{Future Work}

To make this work more useful, we wish to develop a schema to describe the types of edits occurring, similar to the types of schemas exists in other work examining revisions corpora \cite{zhang2017corpus, yang2017identifying, afrin2020annotation}. We are inspired by the Wikipedia Intentions schema developed by \newcite{yang2017identifying}, and are working in collaboration with journalists to further clarify the differences. The development of a news edits schema would help to clarify the nature of these edits as well as focus further questions.

We are also interested in extending prior work \cite{spangher2020sourcefinding, spangher2021multitask} in this domain. We are pursuing collaborations \cite{spangher2021annenberg}. Please do not hesitate to reach the authors by email to discuss any possible collaborations or usages of the dataset.
\FloatBarrier
\bibliographystyle{acl_natbib}
\bibliography{custom}

\begin{thebibliography}{59}
\expandafter\ifx\csname natexlab\endcsname\relax\def\natexlab#1{#1}\fi

\bibitem[{Abujar et~al.(2019)Abujar, Hasan, and Hossain}]{abujar2019sentence}
Sheikh Abujar, Mahmudul Hasan, and Syed~Akhter Hossain. 2019.
\newblock Sentence similarity estimation for text summarization using deep
  learning.
\newblock In \emph{Proceedings of the 2nd International Conference on Data
  Engineering and Communication Technology}, pages 155--164. Springer.

\bibitem[{Achananuparp et~al.(2008)Achananuparp, Hu, and
  Shen}]{achananuparp2008evaluation}
Palakorn Achananuparp, Xiaohua Hu, and Xiajiong Shen. 2008.
\newblock The evaluation of sentence similarity measures.
\newblock In \emph{International Conference on data warehousing and knowledge
  discovery}, pages 305--316. Springer.

\bibitem[{Afrin et~al.(2020)Afrin, Wang, Litman, Matsumura, and
  Correnti}]{afrin2020annotation}
Tazin Afrin, Elaine~Lin Wang, Diane Litman, Lindsay~Clare Matsumura, and
  Richard Correnti. 2020.
\newblock Annotation and classification of evidence and reasoning revisions in
  argumentative writing.
\newblock In \emph{Proceedings of the Fifteenth Workshop on Innovative Use of
  NLP for Building Educational Applications}, pages 75--84.

\bibitem[{Allan et~al.(2003)Allan, Wade, and Bolivar}]{allan2003retrieval}
James Allan, Courtney Wade, and Alvaro Bolivar. 2003.
\newblock Retrieval and novelty detection at the sentence level.
\newblock In \emph{Proceedings of the 26th annual international ACM SIGIR
  conference on Research and development in informaion retrieval}, pages
  314--321.

\bibitem[{Anthonio et~al.(2020)Anthonio, Bhat, and
  Roth}]{anthonio2020wikihowtoimprove}
Talita Anthonio, Irshad Bhat, and Michael Roth. 2020.
\newblock wikihowtoimprove: A resource and analyses on edits in instructional
  texts.
\newblock In \emph{Proceedings of The 12th Language Resources and Evaluation
  Conference}, pages 5721--5729.

\bibitem[{Bhat et~al.(2020)Bhat, Anthonio, and Roth}]{bhat2020towards}
Irshad Bhat, Talita Anthonio, and Michael Roth. 2020.
\newblock \href {https://doi.org/10.18653/v1/2020.emnlp-main.675} {Towards
  modeling revision requirements in wiki{H}ow instructions}.
\newblock In \emph{Proceedings of the 2020 Conference on Empirical Methods in
  Natural Language Processing (EMNLP)}, pages 8407--8414, Online. Association
  for Computational Linguistics.

\bibitem[{Bravo-Marquez and Manriquez(2012)}]{bravo2012zipf}
Felipe Bravo-Marquez and Manuel Manriquez. 2012.
\newblock A zipf-like distant supervision approach for multi-document
  summarization using wikinews articles.
\newblock In \emph{International Symposium on String Processing and Information
  Retrieval}, pages 143--154. Springer.

\bibitem[{Brisbane(2012)}]{brisbane_2012}
Arthur~S. Brisbane. 2012.
\newblock \href
  {https://www.nytimes.com/2012/07/01/opinion/sunday/article-changes-are-shown-in-a-tool-created-by-outsiders.html}
  {Insider's view of changes, from outside}.
\newblock \emph{The New York Times}.

\bibitem[{Burke(2016)}]{burke_2016}
Austin Burke. 2016.
\newblock \href
  {https://www.vrresearch.com/blog/2016/9/1/newsdiffs-a-tool-for-tracking-changes-to-online-news-articles}
  {Newsdiffs: A tool for tracking changes to online news articles - vr research
  - public records research: Opposition research}.

\bibitem[{Carlson et~al.(2003)Carlson, Marcu, and
  Okurowski}]{carlson2003building}
Lynn Carlson, Daniel Marcu, and Mary~Ellen Okurowski. 2003.
\newblock Building a discourse-tagged corpus in the framework of rhetorical
  structure theory.
\newblock In \emph{Current and new directions in discourse and dialogue}, pages
  85--112. Springer.

\bibitem[{Chen et~al.(2018)Chen, Kim, Wilbur, and Lu}]{chen2018sentence}
Qingyu Chen, Sun Kim, W~John Wilbur, and Zhiyong Lu. 2018.
\newblock Sentence similarity measures revisited: ranking sentences in pubmed
  documents.
\newblock In \emph{Proceedings of the 2018 ACM International Conference on
  Bioinformatics, Computational Biology, and Health Informatics}, pages
  531--532.

\bibitem[{Choubey et~al.(2020)Choubey, Lee, Huang, and
  Wang}]{choubey2020discourse}
Prafulla~Kumar Choubey, Aaron Lee, Ruihong Huang, and Lu~Wang. 2020.
\newblock \href {https://doi.org/10.18653/v1/2020.acl-main.478} {Discourse as a
  function of event: Profiling discourse structure in news articles around the
  main event}.
\newblock In \emph{Proceedings of the 58th Annual Meeting of the Association
  for Computational Linguistics}, pages 5374--5386, Online. Association for
  Computational Linguistics.

\bibitem[{Daxenberger and Gurevych(2012)}]{daxenberger2012corpus}
Johannes Daxenberger and Iryna Gurevych. 2012.
\newblock \href {https://www.aclweb.org/anthology/C12-1044} {A corpus-based
  study of edit categories in featured and non-featured {W}ikipedia articles}.
\newblock In \emph{Proceedings of {COLING} 2012}, pages 711--726, Mumbai,
  India. The COLING 2012 Organizing Committee.

\bibitem[{Daxenberger and Gurevych(2013)}]{daxenberger2013automatically}
Johannes Daxenberger and Iryna Gurevych. 2013.
\newblock Automatically classifying edit categories in wikipedia revisions.
\newblock In \emph{Proceedings of the 2013 Conference on Empirical Methods in
  Natural Language Processing}, pages 578--589.

\bibitem[{Faigley and Witte(1981)}]{faigley1981analyzing}
Lester Faigley and Stephen Witte. 1981.
\newblock Analyzing revision.
\newblock \emph{College composition and communication}, 32(4):400--414.

\bibitem[{Faruqui et~al.(2018)Faruqui, Pavlick, Tenney, and
  Das}]{faruqui2018wikiatomicedits}
Manaal Faruqui, Ellie Pavlick, Ian Tenney, and Dipanjan Das. 2018.
\newblock \href {https://doi.org/10.18653/v1/D18-1028} {{W}iki{A}tomic{E}dits:
  A multilingual corpus of {W}ikipedia edits for modeling language and
  discourse}.
\newblock pages 305--315.

\bibitem[{Fass and Main(2014)}]{fass2014revealing}
John Fass and Angus Main. 2014.
\newblock Revealing the news: How online news changes without you noticing.
\newblock \emph{Digital Journalism}, 2(3):366--382.

\bibitem[{Fong and Biuk-Aghai(2010)}]{fong2010did}
Peter Kin-Fong Fong and Robert~P Biuk-Aghai. 2010.
\newblock What did they do? deriving high-level edit histories in wikis.
\newblock In \emph{Proceedings of the 6th International Symposium on Wikis and
  Open Collaboration}, pages 1--10.

\bibitem[{Fu et~al.(2018)Fu, Tan, Peng, Zhao, and Yan}]{fu2018style}
Zhenxin Fu, Xiaoye Tan, Nanyun Peng, Dongyan Zhao, and Rui Yan. 2018.
\newblock Style transfer in text: Exploration and evaluation.
\newblock In \emph{Proceedings of the AAAI Conference on Artificial
  Intelligence}, volume~32.

\bibitem[{Gourarie(2015)}]{gourarie_2015}
Chava Gourarie. 2015.
\newblock \href {https://www.cjr.org/watchdog/newsdiffs_new_york_times.php}
  {Why 'diffing' could make news organizations more transparent}.
\newblock \emph{Columbia Journalism Review}.

\bibitem[{Grundkiewicz and Junczys-Dowmunt(2014)}]{grundkiewicz2014wiked}
Roman Grundkiewicz and Marcin Junczys-Dowmunt. 2014.
\newblock The wiked error corpus: A corpus of corrective wikipedia edits and
  its application to grammatical error correction.
\newblock In \emph{International Conference on Natural Language Processing},
  pages 478--490. Springer.

\bibitem[{Herrmann(2006)}]{bbc2006}
Steve Herrmann. 2006.
\newblock \href
  {https://www.bbc.co.uk/blogs/theeditors/2006/10/sniffing_out_edits.html} {The
  editors: Sniffing out edits}.
\newblock \emph{BBC}.

\bibitem[{Hitomi et~al.(2017)Hitomi, Tamori, Okazaki, and
  Inui}]{hitomi2017proofread}
Yuta Hitomi, Hideaki Tamori, Naoaki Okazaki, and Kentaro Inui. 2017.
\newblock Proofread sentence generation as multi-task learning with editing
  operation prediction.
\newblock In \emph{Proceedings of the Eighth International Joint Conference on
  Natural Language Processing (Volume 2: Short Papers)}, pages 436--441.

\bibitem[{Johnson et~al.(2016)Johnson, Schreiner, and
  Agnone}]{johnson2016effect}
Erik~W Johnson, Jonathan~P Schreiner, and Jon Agnone. 2016.
\newblock The effect of new york times event coding techniques on social
  movement analyses of protest data.
\newblock In \emph{Narratives of Identity in Social Movements, Conflicts and
  Change}. Emerald Group Publishing Limited.

\bibitem[{Jones and Neubert(2017)}]{jones2017using}
Gina~M Jones and Michael Neubert. 2017.
\newblock Using rss to improve web harvest results for news web sites.
\newblock \emph{Journal of Western Archives}, 8(2):3.

\bibitem[{Kajiwara and Komachi(2016)}]{kajiwara2016building}
Tomoyuki Kajiwara and Mamoru Komachi. 2016.
\newblock Building a monolingual parallel corpus for text simplification using
  sentence similarity based on alignment between word embeddings.
\newblock In \emph{Proceedings of COLING 2016, the 26th International
  Conference on Computational Linguistics: Technical Papers}, pages 1147--1158.

\bibitem[{Lan et~al.(2020)Lan, Chen, Goodman, Gimpel, Sharma, and
  Soricut}]{lan2019albert}
Zhenzhong Lan, Mingda Chen, Sebastian Goodman, Kevin Gimpel, Piyush Sharma, and
  Radu Soricut. 2020.
\newblock \href {https://openreview.net/forum?id=H1eA7AEtvS} {{ALBERT:} {A}
  lite {BERT} for self-supervised learning of language representations}.
\newblock In \emph{8th International Conference on Learning Representations,
  {ICLR} 2020, Addis Ababa, Ethiopia, April 26-30, 2020}. OpenReview.net.

\bibitem[{Le et~al.(2018)Le, Wang, Quan, He, and Yao}]{le2018acv}
Yuquan Le, Zhi-Jie Wang, Zhe Quan, Jiawei He, and Bin Yao. 2018.
\newblock Acv-tree: A new method for sentence similarity modeling.
\newblock In \emph{IJCAI}, pages 4137--4143.

\bibitem[{Leacock et~al.(2010)Leacock, Chodorow, Gamon, and
  Tetreault}]{leacock2010automated}
Claudia Leacock, Martin Chodorow, Michael Gamon, and Joel Tetreault. 2010.
\newblock Automated grammatical error detection for language learners.
\newblock \emph{Synthesis lectures on human language technologies},
  3(1):1--134.

\bibitem[{Marcus et~al.(1993)Marcus, Santorini, and
  Marcinkiewicz}]{marcus1993building}
Mitchell Marcus, Beatrice Santorini, and Mary~Ann Marcinkiewicz. 1993.
\newblock Building a large annotated corpus of english: The penn treebank.

\bibitem[{Mehrabi et~al.(2020)Mehrabi, Gowda, Morstatter, Peng, and
  Galstyan}]{mehrabi2020man}
Ninareh Mehrabi, Thamme Gowda, Fred Morstatter, Nanyun Peng, and Aram Galstyan.
  2020.
\newblock Man is to person as woman is to location: Measuring gender bias in
  named entity recognition.
\newblock In \emph{Proceedings of the 31st ACM Conference on Hypertext and
  Social Media}, pages 231--232.

\bibitem[{Minard et~al.(2016)Minard, Speranza, Urizar, Altuna, Van~Erp, Schoen,
  and Van~Son}]{minard2016meantime}
Anne-Lyse Minard, Manuela Speranza, Ruben Urizar, Begona Altuna, Marieke
  Van~Erp, Anneleen Schoen, and Chantal Van~Son. 2016.
\newblock Meantime, the newsreader multilingual event and time corpus.
\newblock In \emph{Proceedings of the Tenth International Conference on
  Language Resources and Evaluation (LREC'16)}, pages 4417--4422.

\bibitem[{Muri{\'c} et~al.(2019)Muri{\'c}, Abeliuk, Lerman, and
  Ferrara}]{muric2019collaboration}
Goran Muri{\'c}, Andres Abeliuk, Kristina Lerman, and Emilio Ferrara. 2019.
\newblock Collaboration drives individual productivity.
\newblock \emph{Proceedings of the ACM on Human-Computer Interaction},
  3(CSCW):1--24.

\bibitem[{Ning et~al.(2018)Ning, Wu, and Roth}]{ning2018multi}
Qiang Ning, Hao Wu, and Dan Roth. 2018.
\newblock A multi-axis annotation scheme for event temporal relations.
\newblock \emph{arXiv preprint arXiv:1804.07828}.

\bibitem[{Pustejovsky et~al.(2003)Pustejovsky, Hanks, Sauri, See, Gaizauskas,
  Setzer, Radev, Sundheim, Day, Ferro et~al.}]{pustejovsky2003timebank}
James Pustejovsky, Patrick Hanks, Roser Sauri, Andrew See, Robert Gaizauskas,
  Andrea Setzer, Dragomir Radev, Beth Sundheim, David Day, Lisa Ferro, et~al.
  2003.
\newblock The timebank corpus.
\newblock In \emph{Corpus linguistics}, volume 2003, page~40. Lancaster, UK.

\bibitem[{Quan et~al.(2019)Quan, Wang, Le, Yao, Li, and
  Yin}]{quan2019efficient}
Zhe Quan, Zhi-Jie Wang, Yuquan Le, Bin Yao, Kenli Li, and Jian Yin. 2019.
\newblock An efficient framework for sentence similarity modeling.
\newblock \emph{IEEE/ACM Transactions on Audio, Speech, and Language
  Processing}, 27(4):853--865.

\bibitem[{Reimers and Gurevych(2019)}]{reimers2019sentence}
Nils Reimers and Iryna Gurevych. 2019.
\newblock Sentence-bert: Sentence embeddings using siamese bert-networks.
\newblock \emph{arXiv preprint arXiv:1908.10084}.

\bibitem[{Roessing(2019)}]{roessing2019wikis}
Thomas Roessing. 2019.
\newblock Wikis and wikinews.
\newblock \emph{The International Encyclopedia of Journalism Studies}, pages
  1--5.

\bibitem[{Shah et~al.(2020)Shah, Schuster, and Barzilay}]{shah2020automatic}
Darsh Shah, Tal Schuster, and Regina Barzilay. 2020.
\newblock Automatic fact-guided sentence modification.
\newblock In \emph{Proceedings of the AAAI Conference on Artificial
  Intelligence}, volume~34, pages 8791--8798.

\bibitem[{Shen et~al.(2017)Shen, Lin, Tu, Zhao, Liu, Sun
  et~al.}]{shen2017recent}
Shi-Qi Shen, Yan-Kai Lin, Cun-Chao Tu, Yu~Zhao, Zhi-Yuan Liu, Mao-Song Sun,
  et~al. 2017.
\newblock Recent advances on neural headline generation.
\newblock \emph{Journal of computer science and technology}, 32(4):768--784.

\bibitem[{Spangher et~al.(2020)Spangher, May, Ferrara, and
  Peng}]{spangher2020sourcefinding}
Alexander Spangher, Jonathan May, Emilio Ferrara, and Nanyun Peng. 2020.
\newblock ``don't quote me on that'': Finding mixtures of sources in news
  articles.
\newblock In \emph{Proceedings of Computation+Journalism Conference}.

\bibitem[{Spangher et~al.(2021{\natexlab{a}})Spangher, May, Shiang, and
  Deng}]{spangher2021multitask}
Alexander Spangher, Jonathan May, Sz-rung Shiang, and Lingjia Deng.
  2021{\natexlab{a}}.
\newblock Multitask learning for class-imbalanced discourse classification.
\newblock \emph{arXiv preprint arXiv:2101.00389}.

\bibitem[{Spangher et~al.(2021{\natexlab{b}})Spangher, Scott, and
  Huang-Isherwood}]{spangher2021annenberg}
Alexander Spangher, Amberg-Lynn Scott, and Ke~Huang-Isherwood.
  2021{\natexlab{b}}.
\newblock \href {https://www.instagram.com/p/CNsUlUflYV0/} {``what's the
  diff?'': Examining news article updates and changing narratives during the
  uss theodore roosevelt coronavirus crisis}.
\newblock In \emph{Annenberg Scymposium}.

\bibitem[{Tamori et~al.(2017)Tamori, Hitomi, Okazaki, and
  Inui}]{tamori2017analyzing}
Hideaki Tamori, Yuta Hitomi, Naoaki Okazaki, and Kentaro Inui. 2017.
\newblock Analyzing the revision logs of a japanese newspaper for article
  quality assessment.
\newblock In \emph{Proceedings of the 2017 EMNLP Workshop: Natural Language
  Processing meets Journalism}, pages 46--50.

\bibitem[{Thorsen(2008)}]{thorsen2008journalistic}
Einar Thorsen. 2008.
\newblock Journalistic objectivity redefined? wikinews and the neutral point of
  view.
\newblock \emph{New Media \& Society}, 10(6):935--954.

\bibitem[{Tian et~al.(2020)Tian, Chakrabarty, Morstatter, and
  Peng}]{tian2020identifying}
Yufei Tian, Tuhin Chakrabarty, Fred Morstatter, and Nanyun Peng. 2020.
\newblock Identifying cultural differences through multi-lingual wikipedia.
\newblock \emph{arXiv preprint arXiv:2004.04938}.

\bibitem[{Walker(2006)}]{ace2005}
et~al. Walker, Christopher. 2006.
\newblock Ace 2005 multilingual training corpus ldc2006t06.
\newblock Philadelphia: Linguistic Data Consortium.

\bibitem[{Wang et~al.(2020)Wang, Matsumura, Correnti, Litman, Zhang, Howe,
  Magooda, and Quintana}]{wang2020erevis}
Elaine~Lin Wang, Lindsay~Clare Matsumura, Richard Correnti, Diane Litman,
  Haoran Zhang, Emily Howe, Ahmed Magooda, and Rafael Quintana. 2020.
\newblock erevis (ing): Students’ revision of text evidence use in an
  automated writing evaluation system.
\newblock \emph{Assessing Writing}, 44:100449.

\bibitem[{Wu et~al.(2020)Wu, Kanoulas, de~Rijke, and Lu}]{wu2020wn}
Chuan Wu, Evangelos Kanoulas, Maarten de~Rijke, and Wei Lu. 2020.
\newblock Wn-salience: A corpus of news articles with entity salience
  annotations.
\newblock In \emph{Proceedings of The 12th Language Resources and Evaluation
  Conference}, pages 2095--2102.

\bibitem[{Yang et~al.(2017)Yang, Halfaker, Kraut, and
  Hovy}]{yang2017identifying}
Diyi Yang, Aaron Halfaker, Robert Kraut, and Eduard Hovy. 2017.
\newblock Identifying semantic edit intentions from revisions in wikipedia.
\newblock In \emph{Proceedings of the 2017 Conference on Empirical Methods in
  Natural Language Processing}, pages 2000--2010.

\bibitem[{Yao et~al.(2018)Yao, Liu, and Zhang}]{yao2018novel}
Haipeng Yao, Huiwen Liu, and Peiying Zhang. 2018.
\newblock A novel sentence similarity model with word embedding based on
  convolutional neural network.
\newblock \emph{Concurrency and Computation: Practice and Experience},
  30(23):e4415.

\bibitem[{Yatskar et~al.(2010)Yatskar, Pang, Danescu-Niculescu-Mizil, and
  Lee}]{yatskar2010sake}
Mark Yatskar, Bo~Pang, Cristian Danescu-Niculescu-Mizil, and Lillian Lee. 2010.
\newblock For the sake of simplicity: Unsupervised extraction of lexical
  simplifications from wikipedia.
\newblock \emph{arXiv preprint arXiv:1008.1986}.

\bibitem[{Yimam et~al.(2017)Yimam, {\v{S}}tajner, Riedl, and
  Biemann}]{yimam2017cwig3g2}
Seid~Muhie Yimam, Sanja {\v{S}}tajner, Martin Riedl, and Chris Biemann. 2017.
\newblock Cwig3g2-complex word identification task across three text genres and
  two user groups.
\newblock In \emph{Proceedings of the Eighth International Joint Conference on
  Natural Language Processing (Volume 2: Short Papers)}, pages 401--407.

\bibitem[{Yin et~al.(2018)Yin, Neubig, Allamanis, Brockschmidt, and
  Gaunt}]{yin2018learning}
Pengcheng Yin, Graham Neubig, Miltiadis Allamanis, Marc Brockschmidt, and
  Alexander~L Gaunt. 2018.
\newblock Learning to represent edits.
\newblock \emph{arXiv preprint arXiv:1810.13337}.

\bibitem[{Zanzotto and Pennacchiotti(2010)}]{zanzotto2010expanding}
Fabio~Massimo Zanzotto and Marco Pennacchiotti. 2010.
\newblock Expanding textual entailment corpora fromwikipedia using co-training.
\newblock In \emph{Proceedings of the 2nd Workshop on The People’s Web Meets
  NLP: Collaboratively Constructed Semantic Resources}, pages 28--36.

\bibitem[{Zhang et~al.(2017)Zhang, Hashemi, Hwa, and Litman}]{zhang2017corpus}
Fan Zhang, Homa~B Hashemi, Rebecca Hwa, and Diane Litman. 2017.
\newblock A corpus of annotated revisions for studying argumentative writing.
\newblock In \emph{Proceedings of the 55th Annual Meeting of the Association
  for Computational Linguistics (Volume 1: Long Papers)}, pages 1568--1578.

\bibitem[{Zhang and Litman(2015)}]{zhang2015annotation}
Fan Zhang and Diane Litman. 2015.
\newblock Annotation and classification of argumentative writing revisions.
\newblock \emph{Grantee Submission}.

\bibitem[{Zhang and Wan(2017)}]{zhang2017towards}
Jianmin Zhang and Xiaojun Wan. 2017.
\newblock Towards automatic construction of news overview articles by news
  synthesis.
\newblock In \emph{Proceedings of the 2017 Conference on Empirical Methods in
  Natural Language Processing}, pages 2111--2116.

\bibitem[{Zhang(2020)}]{zhang2020engaging}
Zhe~Victor Zhang. 2020.
\newblock Engaging with automated writing evaluation (awe) feedback on l2
  writing: Student perceptions and revisions.
\newblock \emph{Assessing Writing}, 43:100439.

\end{thebibliography}
\end{document}